\documentclass[11pt,a4paper]{article}
\usepackage[hyperref]{eacl2021}
\usepackage{times}
\usepackage{latexsym}

\usepackage{url}
\usepackage[utf8]{inputenc}

\usepackage{covington}
\usepackage{booktabs}
\usepackage{relsize}
\usepackage{todonotes}
\usepackage{xspace}
\usepackage{amsmath}
\usepackage{hyperref}
\usepackage{multirow}
\usepackage[all]{nowidow}
\usepackage{breakurl}
\usepackage{comment}
\usepackage{tikz-dependency}

\usepackage[page, title]{appendix}

\newcommand\F{{$\text{F}_1$}\xspace}

\newcommand{\ie}{\textit{i.e.,}\xspace}
\newcommand{\eg}{\textit{e.g.,}\xspace}
\newcommand{\f}[1]{\footnotesize{(#1)}} 

\definecolor{lightblue}{RGB}{205,210,255}
\definecolor{lightred}{RGB}{255,240,240}
\newcommand{\imp}[1]{\colorbox{lightblue}{#1}}
\newcommand{\loss}[1]{\colorbox{lightred}{#1}}

\newcommand{\cls}{\textsc{[Cls]}\xspace}
\newcommand{\first}{\textsc{First}\xspace}
\newcommand{\mean}{\textsc{Mean}\xspace}
\newcommand{\maxm}{\textsc{Max}\xspace}
\newcommand{\maxmm}{\textsc{MaxMM}\xspace}

\newcommand{\targets}{\textsc{Targ.}\xspace}
\newcommand{\pall}{\textsc{Pred.}\xspace}
\newcommand{\polar}{\textsc{+Pol.}\xspace}

\aclfinalcopy 
\def\aclpaperid{36} 

\title{If you've got it, flaunt it: \\
       Making the most of fine-grained sentiment annotations}

\author{Jeremy Barnes, 
  Lilja {\O}vrelid, and  
  Erik Velldal \\
  University of Oslo\\
  Department of Informatics\\
 {\tt \{jeremycb,liljao,erikve\}@ifi.uio.no}}

\date{}

\begin{document}
\maketitle
\begin{abstract}
  \textit{Fine-grained} sentiment analysis attempts to extract sentiment holders, targets and polar expressions and resolve the relationship between them, but progress has been hampered by the difficulty of annotation. \textit{Targeted} sentiment analysis, on the other hand, is a more narrow task, focusing on extracting sentiment targets and classifying their polarity.

In this paper, we explore whether incorporating holder and expression information can improve target extraction and classification and perform experiments on eight English datasets.
We conclude that jointly predicting target and polarity BIO labels improves target extraction, and that augmenting the input text with gold expressions generally improves targeted polarity classification. This highlights the potential importance of annotating expressions for fine-grained sentiment datasets. At the same time, our results show that performance of current models for predicting polar expressions is poor, hampering the benefit of this information in practice.

  
  
\end{abstract}

\section{Introduction}
\label{sec:intro}

\begin{figure*}[h!]
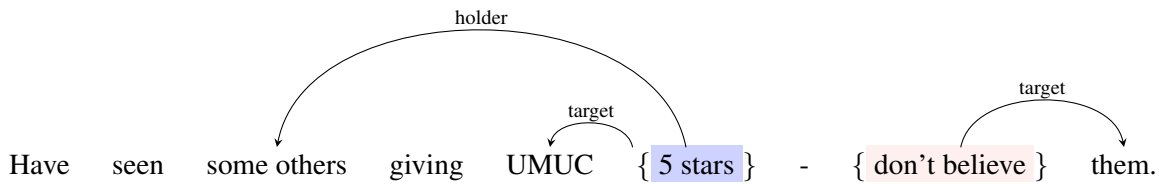

\centering
\normalsize
\begin{dependency}[theme = simple]
   \begin{deptext}[column sep=1em]
    Have \& seen \& some others \& giving \& UMUC \& \{\colorbox{lightblue}{5 stars}\} \& - \& \{\colorbox{lightred}{don't believe}\} \& them.\\
   \end{deptext}
   \depedge[edge start x offset=-20pt]{6}{5}{\normalsize{target}}
   \depedge{6}{3}{\normalsize{holder}}
   \depedge{8}{9}{\normalsize{target}}

\end{dependency}
\caption{An opinion annotation from the Darmstadt Review Corpus.}
\label{deptree}
\end{figure*}

Sentiment analysis comes in many flavors, arguably the most complete of which is what is often called fine-grained sentiment analysis \cite{Wiebe2005,Liu:15}. This approach models the sentiment task as minimally extracting all opinion holders, targets, and expressions in a text and resolving the relationships between them. This complex task is further complicated by interactions between these elements, strong domain effects, and the subjective nature of sentiment. Take the annotated sentence in Figure \ref{deptree} as an example.
Knowing that the target ``UMUC'' is modified by the expression ``5 stars'' and not ``don't believe'' is important to correctly classifying the polarity. Additionally, the fact that this is a belief held by ``some others'' as apposed to the author of the sentence can help us determine the overall polarity expressed in the sentence.

Compared to document- or sentence-level sentiment analysis, where distant labelling schemes can be used to obtain annotated data, fine-grained annotation of sentiment does not occur naturally, which means that current machine learning models are often hampered by the small size of datasets. Furthermore, fine-grained annotation is demanding, leads to relatively small datasets, and has low inter-annotator agreement \cite{Wiebe2005,wang-etal-2017-tdparse}. This begs the question: \textbf{is it worth it to annotate full fine-grained sentiment}?

Targeted sentiment  \cite{mitchell-etal-2013-open,zhang-etal-2015-neural} is a reduction of the fine-grained sentiment task which concentrates on extracting sentiment targets and classifying their polarity, effectively ignoring sentiment holders and expressions. The benefit of this setup is that it is faster to annotate and simpler to model. But \textbf{would targeted sentiment models benefit from knowing the sentiment holders and expressions}?

In this work, we attempt to determine whether holder and expression information is useful for extracting and then classifying sentiment targets. Specifically, we ask the following research questions:

\begin{enumerate}
    \item[] \textbf{RQ1:} Given the time and difficulty required to annotate opinion holders, expressions, and polarity, is this information useful to \textbf{extract sentiment targets}?
    \begin{enumerate}
        \item Does augmenting the input text with holders and expressions improve target extraction?
        \item Do target extraction models benefit from predicting holders and expressions?
        \item Do target extraction models benefit from predicting the polarity of targets and/or expressions?
    \end{enumerate}
    \item[] \textbf{RQ2:} Can holder and expression information improve \textbf{polarity classification} on extracted targets?
    \begin{enumerate}
        \item Does augmenting the input text with holders and expressions improve polarity classification?
        \item Do potential benefits of augmenting the input depend on how we model the target, \ie using the [CLS] embeddings, mean pooling the target embeddings, etc.?
        \item Can sentiment lexicons provide enough information on expressions to give improvements?
    \end{enumerate}
\end{enumerate}

We conduct a series of experiments on eight English sentiment datasets (three with full fine-grained sentiment and five targeted) with state-of-the-art models based on fine-tuned BERT models. We show that (1) it is possible to improve target extraction by also trying to predict the polarity, and that (2) classification models benefit from having access to information about sentiment expressions. We also (3) release the code\footnote{\url{https://github.com/ltgoslo/finegrained_modelling}} 
to reproduce the experiments, as well as the scripts to download, preprocess, and collect the datasets into a compatible JSON format, with the hope that this allows future research on the same data.

\section{Related work}
\label{sec:background}

\textbf{Fine-grained} approaches to sentiment analysis attempt to discover opinions from text, where each opinion is a tuple of \textit{(opinion holder, opinion target, opinion expression, polarity, intensity)}. Annotation of datasets for this granularity requires creating in-depth annotation guidelines, training annotators, and generally leads to lower inter-annotator scores than other sentiment tasks, \eg document- or sentence-level classification, as deciding on the spans for multiple elements and their relationships is undeniably harder than choosing a single label for a full text. \textbf{Targeted sentiment}, on the other hand,
generally concentrates only on target extraction and polarity classification.
This has the benefit of allowing non-experts and crowd-sourcing to perform annotation, making it easier to collect larger datasets for machine learning.
This simplified annotation can be crowd-sourced, leading to larger datasets for machine learning.
\subsection{Datasets}
The Multi-purpose Question Answering dataset (\textit{MPQA}) \cite{Wiebe2005} is the first dataset that annotated opinion holders, targets, expressions and their relationships. The news wire data leads to complex opinions and a generally difficult task for sentiment models. Normally, the full opinion extraction task is modelled as extraction of the individual elements (holders, targets, and expressions) and the subsequent resolution of the relationship between them.

The \textit{Darmstadt Review Corpora} \cite{toprak-etal-2010-sentence} contain annotated opinions for consumer reviews of universities and services. The authors annotate holders, targets, expressions, polarity, modifiers, and intensity. They achieve between 0.5 and 0.8 agreement using the $agr$ method \cite{Wiebe2005}, with higher disagreement on what they call ``polar targets'' -- targets that have a polarity but no annotated sentiment expression -- holders, and expressions.

The \textit{Open} Domain Targeted dataset \cite{mitchell-etal-2013-open} makes use of crowd sourcing to annotate NEs from scraped tweets in English and Spanish \cite{nerit-named-entity-recognition-for-informal-text} with their polarities. The authors use majority voting to assign the final labels for the NEs, discarding tweets without sentiment consensus on all NEs. 

The 2014 \textit{SemEval} shared task \cite{pontiki-etal-2014-semeval} on aspect-based sentiment analysis include labeled data from restaurant and laptop reviews for two subtasks: 1) target extraction, which they call ``aspect term extraction'' and 2) classification of polarity with respect to targets (``aspect term polarity''). 

As most targeted datasets only contain a single target, or multiple targets with the same polarity, sentence-level classifiers are strong baselines. In order to mitigate this, \newcite{jiang-etal-2019-challenge} create a \textit{Challenge} dataset which has both multiple targets and multiple polarities in each sentence. 
%
Similarly, \newcite{wang-etal-2017-tdparse} also point out that most targeted sentiment methods perform poorly with multiple targets and propose \textit{TDParse}, a corpus of UK election tweets with multiple targets per tweet. 
%

\subsection{Modelling}

\newcite{katiyar-cardie-2016-investigating} explore jointly extracting holders, targets, and expressions with LSTMs. They find that adding sentence-level and relation-level dependencies (\textsc{is-from} or \textsc{is-about}) improve extraction, but find that the LSTM models lag behind CRFs with rich features.

\begin{table*}[t]
\newcommand{\sep}{\cmidrule(lr){3-3}\cmidrule(lr){5-6}\cmidrule(lr){7-9}\cmidrule(lr){10-12}\cmidrule(lr){13-15}\cmidrule(lr){16-18}}
    \centering
    \resizebox{\textwidth}{!}{%
    \begin{tabular}{llllrrrrrrrrrrrrrrrrrr}
    \toprule
      & & domain & &\multicolumn{2}{c}{sentences} & \multicolumn{3}{c}{holders} & \multicolumn{3}{c}{targets} & \multicolumn{3}{c}{expressions} & \multicolumn{3}{c}{polarity}\\
      \sep
      & & & &\# & avg. & \# & avg. & max & \# & avg. & max & \# & avg. & max & $+$ & neu & $-$\\
    \sep
  \multirow{9}{*}{\rotatebox{90}{Fine-grained Sentiment}}  &  MPQA & newswire      & train & 4500 & 25 & 1306 & 2.6 & 27  & 1382 & 6.1 & 56 &  1656 & 2.4 & 14  & 675 & 271 & 658 \\
               & &  & dev & 1622 & 23 & 377 & 2.6 & 16 & 449 & 5.3 & 41 & 552 & 2.1 & 8 & 241 & 105 & 202\\
                 & & & test & 1681 & 24 & 371 & 2.8 & 32 & 405 & 6.4 & 42 & 479 & 2.0 & 8 & 166 & 89 & 199\\
      \sep
      & DS. Services & service & train & 5913 & 16 & 18 & 1.2 & 2 & 2504 & 1.2 & 7 & 1273 & 1.2 & 10 & 1623 & 46 & 838\\
                & & reviews & dev & 744 & 18 & 1 & 1.7 & 3 & 288 & 1.2 & 4 & 144 & 1.4 & 5  & 103 & 1 & 104\\
                & & & test & 748 & 17 & 2 & 1 & 1 & 328 & 1.2 & 5 & 168 & 1.4 & 6 & 241 & 7 & 80\\
      \sep
      & DS. Uni & university &  train & 2253 & 20 & 65 & 1.2 & 2 & 1252 & 1.2 & 5 & 837 & 1.9 & 9 & 495 & 149 & 610 \\
               & & reviews & dev & 232 & 9 & 17 & 1.1 & 3 & 151 & 1.2 & 3 & 106 & 1.7 & 6  & 40 & 19 & 92\\
                & & & test & 318 & 20 & 12 & 1.3 & 4 & 198 & 1.2 & 6 & 139 & 2.0 & 5 &  77 & 18 & 103\\
                 
    \midrule
    
    \multirow{15}{*}{\rotatebox{90}{Targeted Sentiment}}  &  TDParse & political & train & 2889 & 6.9 & - & - & - & 9088 & 1.2 & 7 & - & - & - & 1238 & 3931 & 3919  \\
         &   & tweets  & dev & 321 & 6.6 & - & - & - & 1040 & 1.2 & 5 & - & - & - &  128 & 454 & 458\\
          &  &   & test & 867 & 6.9 & - & - & - & 2746 & 1.2 & 6 & - & - & -& 378 & 1162 & 1206 \\
            \sep
      & SemEval R. & restaurant & train & 2740 & 13 & - & -  & - & 3293 & 1.4 & 19 & - & - & - & 1902 & 574 & 734\\
              &   &  reviews  &  dev & 304 & 11.3 & - & - & - & 350 & 1.4 & 5 &  - & - & -& 226 & 54 & 63\\
              &   &    &  test & 800 & 9.6& - & - & - &  1128 & 1.4 & 8 &  - & - & -& 724 & 195 & 195\\
                 \sep
    & SemEval L. & laptop & train & 2744 & 22.5 & - & - & - & 2049 & 1.5 & 6  & -& - & - & 870 & 402 & 747 \\
              &   &  reviews  &  dev & 304 & 21.1 & - & - & -  & 244 & 1.6 & 5 & - & - &- & 99 & 44 & 96\\
             &    &    &  test & 800 & 18.6 & - & - & -  & 633 & 1.6 & 7 & - & - & -& 327 & 162 & 128   \\
                 \sep
     &  Open & tweets & train & 1903 & 12.8  & - & - & - & 2594 & 1.6 & 8 & - & - & - & 578 & 1801 & 215\\
             &   &          & dev & 211 & 12.3  & - & - & - & 291  & 1.6 & 6 & - & - & - & 46 & 220 & 25 \\
             &   &          & test & 234 & 11.6  & - & - & - & 337 & 1.6 & 7 & - & - & - & 74 & 232 & 31 \\
                \sep
     &  Challenge & restaurant &train & 4297 & 8.8  & - & - & - & 11186 & 1.3 & 9  & - & - & -& 3380 & 5042 & 2764\\
            &  & reviews & dev &  500 & 8.9  & - & - & - & 1332 & 1.3 & 8 &  - & - & - & 403 & 604 & 325\\
           &   &  & test & 500 & 8.9  & - & -  & - & 1336 & 1.3 & 8  & - & - & - & 400 & 607 & 329\\
      \bottomrule
    \end{tabular}
}%
    
    \caption{Stastistics of the datasets, including number of sentences, as well as average, and max lengths (in tokens) for holder, target, and expression annotations. Additionally, we include the distribution of polarity -- restricted to positive, neutral, and negative -- in each dataset.}
    \label{tab:statistics}
\end{table*}

Regarding modelling the interaction between elements, there are several previous attempts to jointly learn to extract and classify targets, using factor graphs \cite{klinger-cimiano-2013-bi}, multi-task learning \cite{he-etal-2019-interactive} or sequence tagging with collapsed tagsets representing both tasks \cite{li-etal-2019-unified}. In general, the benefits are small and have suggested that there is only a weak relationship between target extraction and polarity classification \cite{hu-etal-2019-open}.

\section{Data}
\label{sec:data}

One of the difficulties of working with fine-grained sentiment analysis is that there are only a few datasets (even in English) and they come in incompatible, competing data formats, \eg BRAT or various flavors of XML. With the goal of creating a simple unified format to work on fine-grained sentiment tasks, we take the eight datasets mentioned in Section \ref{sec:background} -- \textit{MPQA} \cite{Wiebe2005}, \textit{Darmstadt Services and Universities} \cite{toprak-etal-2010-sentence}, \textit{TDParse} \cite{wang-etal-2017-tdparse}, \textit{SemEval Restaurant and Laptop} \cite{pontiki-etal-2014-semeval}, \textit{Open} Domain Targeted Sentiment \cite{mitchell-etal-2013-open}, and the \textit{Challenge} dataset from \newcite{jiang-etal-2019-challenge} -- and convert them to a standard \emph{JSON} format. The datasets are sentence and word tokenized using NLTK \cite{Loper2002}, except for MPQA, DS. Service and DS. Uni, which already contain sentence and token spans. All polarity annotations are mapped to positive, negative, neutral, and conflict\footnote{We discard conflict during evaluation because there are not enough examples to properly learn this class in most datasets}. As such, each sentence contains a sentence id, the tokenized text, and a possibly empty set of opinions which contain a holder, target, expression, polarity, and intensity. We allow for empty holders and expressions in order generalize to the targeted corpora. Finally, we use 10 percent of the training data as development and another 10 percent for test for the corpora that do not contain a suggested train/dev/test split. For training and testing models, however, we convert the datasets to CoNLL format. 

Table \ref{tab:statistics} presents an overview of the different datasets and highlights important differences between them. The fully fine-grained sentiment datasets (MPQA, DS. Services, and DS. Uni) tend to be larger but have fewer targets annotated, due to a larger number of sentences with no targets. However, the MPQA dataset contains much longer targets than the other datasets -- an average of 6, but a maximum of 56 tokens. It also contains more opinion holders and expressions and these also tend to be longer, all of which marks MPQA as an outlier among the datasets. The distribution of polarity is also highly dependent on the dataset, with DS. Services being the most skewed and SemEval Laptop the least skewed. Finally, the challenge dataset is by far the largest with over 11,000 training targets. Additionally, Table \ref{tab:targets} in Appendix \ref{app:additional} shows the percentage of unique targets per dataset, as well as the percentage of targets shared between the training set and the dev and test sets. Again MPQA has the largest number of unique targets and the least overlap.\footnote{We do not, however, consider partial overlap which may exaggerate the true uniqueness of targets.} 

\section{Experimental Setup}
\label{sec:models}

\begin{figure*}
    \centering
    \includegraphics[width=.8\textwidth]{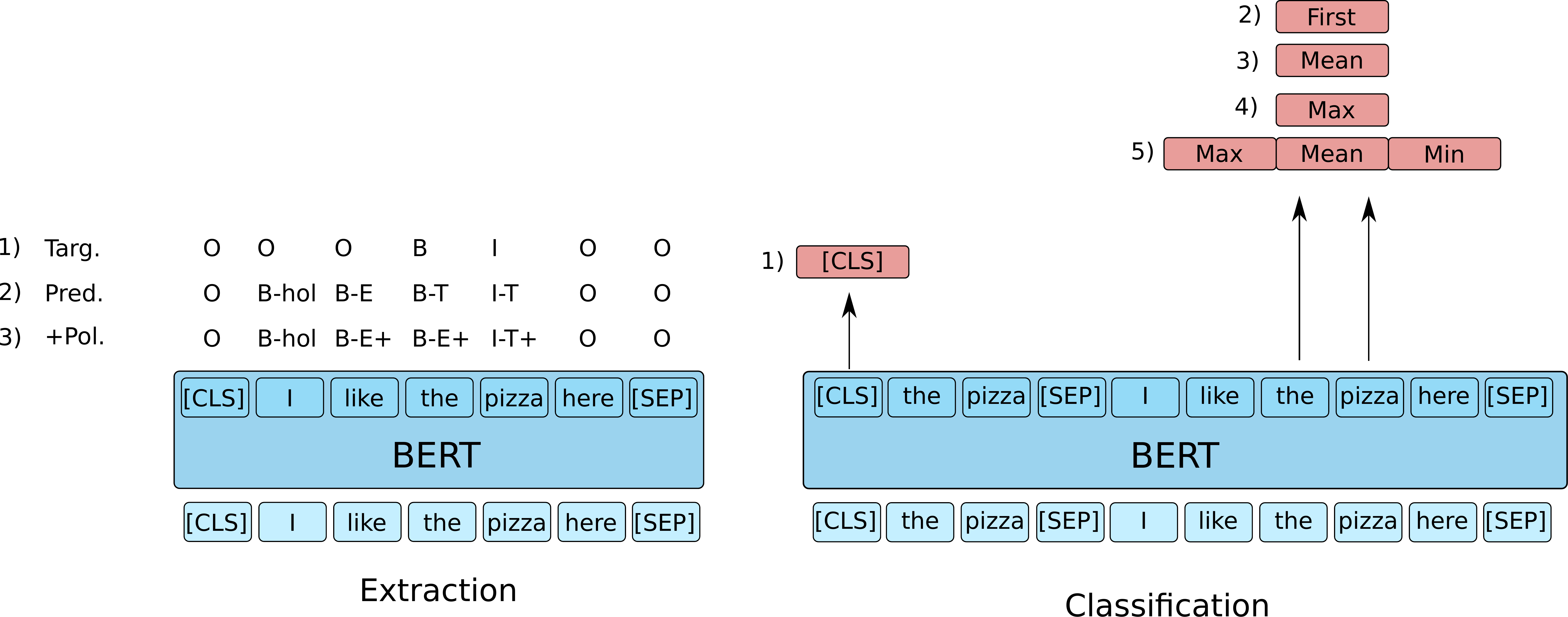}
    \caption{Our BERT-based \emph{target extraction} and \emph{classification models}, with the three strategies for extraction ((1) predict only targets, (2) predict holders, targets and expressions, and (3) predict the polarity of the targets and expressions as well) and five strategies for sentiment classification (passing to the softmax layer the contextualized embedding from (1) the [CLS] embedding, (2) the first token in the target (3) averaging all embeddings in the target phrase, (4) taking the max of the target embeddings, (5) concatenating the max, mean, and min).}
    \label{fig:models}
\end{figure*}

We split the task of 
targeted sentiment analysis 
into the \textbf{extraction of sentiment targets} and subsequent \textbf{polarity classification} of extracted targets, given their context. Figure \ref{fig:models} shows the two tasks and the eight models used in the experiments. As a base model, we take the target extraction and classification models from \newcite{xu-etal-2019-bert}, which achieve state-of-the-art performance on the SemEval task. The approach first fine-tunes BERT \cite{devlin-etal-2019-bert} on domain-specific unlabeled data as a domain-adaptation step. We use the datasets themselves to perform this step, except for the SemEval datasets. For these, we follow \newcite{rietzler-etal-2020-adapt} and instead use larger amounts of unlabeled data -- 1,710,553 and 2,000,000 sentences for SemEval Laptop and Restaurant respectively -- taken from Amazon Laptop reviews \cite{He2016-ups-downs} and the Yelp Dataset Challenge.\footnote{\url{https://www.yelp.com/dataset/challenge}} We further deviate from \newcite{xu-etal-2019-bert} by not pretraining the models on the SQUAD question answering dataset and in-domain sentiment questions which they create, as this data is not publicly available. Finally, a linear prediction is added after the BERT model and the full model is updated on the sentiment task.

For \textbf{target extraction}, we use the contextualized BERT embeddings as input to a softmax layer and predict the sequence of tags. We compare three prediction strategies:

\begin{enumerate}\itemsep-2pt
    \item \textbf{\targets:} The model predicts the labels $y \in \{\textit{B,I,O}\}$ for the targets only.
    \item \textbf{\pall:} We additionally predict the labels for holders and expressions and predict $y \in \{\textit{B-holder},\textit{I-holder},\textit{B-target},\textit{I-target},$
    $\textit{B-expression},\textit{I-expression},O\}$.
    \item \textbf{\polar:} Finally, we add the polarity  (\textit{positive}, \textit{negative}, \textit{neutral}) to the annotation specific BIO-tag, which leads to an inventory of 19 labels for the full fine-grained setup and 7 for the targeted setup.
\end{enumerate}

For \textbf{polarity classification}, we take as a baseline the classification architecture from \newcite{xu-etal-2019-bert}, which makes use of the two-sentence training procedure for BERT, by prepending the target before the sentence separation token, and then adding the full sentence after. We compare five strategies for producing the input to the softmax layer for predicting the sentiment of the target:

\begin{enumerate}\itemsep-2pt
    \item \textbf{\cls:} this model uses the [CLS] embedding from the final BERT layer.  
    \item \textbf{\first:} uses the contextualized BERT embedding from the first token of the target in context.  
    \item \textbf{\mean:} instead takes the average of the BERT embeddings for the tokens in the target. 
    \item \textbf{\maxm:}  uses the max of the contextualized BERT embeddings for the tokens in the target.

    \item \textbf{\maxmm:} takes the max, min, and mean pooled representations and passes the concatenation to the softmax layer, which has shown to perform well for sentiment tasks \cite{tang-etal-2014-learning}. However, this triples the size of the input representation to the softmax layer.

\end{enumerate}

The \targets and \cls models correspond to the models used in \newcite{xu-etal-2019-bert} and serve as baselines. The extraction and classification models are fine-tuned for 50 epochs using Adam with an initial learning rate of $3\mathrm{e}{-5}$, with a linear warmup of 0.1 and all other hyperparameters are left at default BERT settings (further details in Appendix \ref{app:trainingdetails}). The best model on the development set is used for testing. Combined with the four input manipulations (Table \ref{tab:insert_tags}), this leads to eleven extraction experiments -- \targets and \pall on the original data which only has annotated targets are the same and for simplicity we only show the results from \targets -- and twenty classification experiments per dataset. In order to control for the effect of random initialization, we run each experiment 5 times on different random seeds and report the mean and standard deviation.

\subsection{Training with gold annotations}

\begin{table*}[]
\definecolor{lightgreen}{RGB}{210,240,215}
\definecolor{lightgrey}{RGB}{200,200,200}
\definecolor{lighty}{RGB}{225,240,240}
\newcommand{\targ}[1]{\scriptsize{\colorbox{lightgreen}{\big[\textless T \big]}} \normalsize{#1}  \scriptsize{\colorbox{lightgreen}{\big[T\textgreater \big]}} \normalsize }
\newcommand{\hol}[1]{\scriptsize{\colorbox{lightgrey}{\big[\textless H \big]}} \normalsize{#1}  \scriptsize{\colorbox{lightgrey}{\big[H\textgreater \big]}} \normalsize}
\newcommand{\expr}[1]{\scriptsize{\colorbox{lighty}{\big[\textless E \big]}} \normalsize{#1}  \scriptsize{\colorbox{lighty}{\big[E\textgreater \big]}} \normalsize}
    \centering
     \resizebox{\textwidth}{!}{%
    \begin{tabular}{ll}
\toprule
       original  & Money Magazine rated E-Trade highly . \\
 + holders     & \hol{Money Magazine} rated E-Trade highly \\
   + expressions & Money Magazine \expr{rated} E-Trade \expr{highly}\\
    + full &  \hol{Money Magazine} \expr{rated} E-Trade \expr{highly}\\
\bottomrule
    \end{tabular}
    }
    \caption{We inform our models regarding annotations other than targets by inserting special tags into the input text before and after annotated \colorbox{lightgrey}{holders} and \colorbox{lighty}{expressions}.}
    \label{tab:insert_tags}
\end{table*}

Given that we are interested in knowing whether it is beneficial to include information about additional annotations (holder, expressions, polarity), we perform experiments where we systematically include these. We do so by adding special tags, \eg, \big[\textless E \big], into the input text surrounding the annotated spans, as shown in Table \ref{tab:insert_tags}. The models then have access to this information both during training and at test time, albeit in an indirect way. For the first set of experiments, we perform controlled experiments under ideal conditions, \ie having gold annotations during testing. This allows us to isolate the effects of incorporating the additional annotations, without worrying about noisy predictions

\subsection{Training with predicted expressions}

It is equally important to know whether the models are able to use noisy predicted annotations. In order to test this, we train \emph{expression prediction models} on the three full fine-grained sentiment corpora. We use the same BERT-based model and hyperparameters from the target extraction models above 
and train five models with different random seeds. Preliminary results suggested that these models had high precision, but low recall. Therefore, we take a simple ensemble of the five trained models, where for each token, we keep labels predicted by at least one of the expression models in order to increase recall.

We perform an additional set of experiments where we use sentiment lexicons and assume any word in these lexicons is a sentiment expression. We use the \textbf{Hu and Liu} lexicon \cite{hu2004mining}, the \textbf{SoCal} and \textbf{SoCal-Google} lexicons \cite{taboada-etal-2006-methods} and the \textbf{NRC} emotion lexicon \cite{Mohammad13}, which also contains sentiment annotations. The lexicons contain 6,789, 5,824, 2,142, and 5,474 entries, respectively.
The MPQA and Darmstadt experiments show the effect of predicted vs. gold expressions, as well as domain transfer. The experiments on the targeted datasets, on the other hand, will show us whether it is possible to improve the targeted models with predicted expressions.

\section{Results}

\begin{table*}[]
\newcommand{\sep}{\cmidrule(lr){3-3}\cmidrule(lr){4-4}\cmidrule(lr){5-5}\cmidrule(lr){6-6}\cmidrule(lr){7-7}\cmidrule(lr){8-8}\cmidrule(lr){9-9}\cmidrule(lr){10-10}}
    \centering
    \resizebox{\textwidth}{!}{%
    \begin{tabular}{llrrrrrrrrr}
    \toprule
  &  & MPQA & DS. Services & DS. Unis & Challenge & SemEval R. & SemEval L. & Open & TDParse\\
    \sep
& \newcite{xu-etal-2019-bert} & n/a & n/a &n/a & n/a& 78.0 & 84.3 & n/a & n/a \\
& BiLSTM-CRF & 12.2 \f{1} & 85.0 \f{1} &84.4 \f{1} & 73.4 \f{1} & 72.5 \f{1} & 74.0 \f{1} & 62.2 \f{1} & 82.6\\
\midrule

& original &  14.1 \f{2} &  85.9 \f{1} &  \textbf{84.6 \f{0}} &  75.8 \f{1} &   51.9 \f{1} & 71.3 \f{1} & 62.0 \f{4} &    81.7 \f{3} \\
\sep

\multirow{3}{*}{\rotatebox{90}{\targets}}
 &+ holders &  \loss{11.9 \f{1}} & \loss{84.3 \f{1}} & \loss{83.6 \f{1}} &  - & - & - & - & -\\
 &+ exp. &  \loss{11.6 \f{1}} & \loss{85.0 \f{0}} & \loss{83.4 \f{0}} &  - & - & - & - & -\\
 &+ full &  \loss{10.5 \f{2}} & \loss{84.8 \f{1}} & \loss{83.8 \f{1}} &  - & - & - & - & -\\

\sep

\multirow{4}{*}{\rotatebox{90}{\pall}}
 & + holders  &  \loss{12.1 \f{2}} &  \imp{\textbf{86.2 \f{0}}} &  \textbf{84.6 \f{0}} &  - & - & - & - & -\\
& + exp.  &  \imp{\textbf{14.9 \f{1}}} &  \loss{84.7 \f{1}} &  84.5 \f{1} & - & - & - & - & -\\
 & + full    &  13.0 \f{3} &  \loss{85.5 \f{1}} &  \loss{84.3 \f{1}} &  - & - & - & - & -\\
 
\sep
 
\multirow{6}{*}{\rotatebox{90}{\polar}}
& BiLSTM-CRF & \loss{13.9 \f{1}} & \imp{85.2 \f{1}} & \loss{83.7 \f{1}} & \imp{73.6 \f{1}} & \imp{\textbf{73.7 \f{1}}} & \imp{\textbf{74.5 \f{1}}} &   \imp{62.3 \f{1}} & \loss{81.8 \f{1}} \\
& original     &  \imp{13.8 \f{1}} &  \loss{85.4 \f{1}} &  \loss{84.3 \f{1}} &  \imp{\textbf{76.9 \f{1}}} &   \imp{52.5 \f{1}} &    \imp{\textbf{71.6 \f{1}}} &    \imp{\textbf{62.9 \f{1}}} &    \imp{\textbf{83.2 \f{0}}} \\

\sep

& + holders    & \imp{ 13.8 \f{2}} &  \loss{85.6 \f{1}} &  84.4 \f{1} & - & - & - & - & -\\
& + exp. &  \imp{13.5 \f{2}} &  \loss{85.4 \f{1}} & \loss{ 84.3 \f{0}} &  - & - & - & - & -\\
 & + full                &  \loss{12.0 \f{1}} &  86.0 \f{1} &  \textbf{84.6 \f{0}}    &  - & - & - & - & -\\

       \bottomrule
    \end{tabular}
}%
    
    \caption{Average token-level F1 scores for the \underline{target extraction task} across five runs, (standard deviation in parenthesis). \textbf{Bold} numbers indicate the best model per dataset, while \imp{blue} and \loss{pink} highlighting indicates an \imp{improvement} or \loss{loss in performance} compared to the original data, respectively.}
    \label{tab:extractionresults}
\end{table*}

In this section we describe the main results from the extraction and two classification experiments described in Section \ref{sec:models}.

\subsection{Target extraction}

Table \ref{tab:extractionresults} shows the results for the extraction experiment, where token-level \F is measured only on targets. The models perform poorer than the state-of-the-art, as we did not finetune on the SQUAD question answering dataset and in-domain sentiment questions or perform extensive hyperparameter tuning. The average \F score depends highly on the dataset -- MPQA is the most difficult dataset with 13.1 \F on the original data,  while the Darmstadt Universities corpus is the easiest for target extraction with 84.6. Augmenting the input text with further annotations, but predicting only sentiment targets (\targets in Table \ref{tab:extractionresults}) hurts the model performance in all cases. Specifically, adding holder tags leads to an average drop of 1.3 percentage points (pp), expressions 1.2 and full 1.5. Attempting to additionally predict these annotations (\pall in Table \ref{tab:extractionresults}) leads to mixed results -- the model leads to improvements on MPQA + exp. and Darmstadt Services + holders, no notable difference on MPQA + full and Darmstadt Universities + exp., and a loss on the rest.

Adding the polarity to the target BIO tags (original \polar in Table \ref{tab:extractionresults}) leads to the most consistent improvements across experiments -- an average of 0.5 pp -- with the largest improvement of 1.5 pp on the TDParse dataset. This suggests a weak-to-moderate relationship between polarity and extraction, which contradicts previous conclusions \cite{hu-etal-2019-open}. 
Finally, further adding the holder and expression tags (\polar in Table \ref{tab:extractionresults}) tends to decrease performance.

\subsection{Polarity classification with gold annotations}
Table \ref{tab:sentimentresults} shows the macro \F scores for the polarity classification task on the gold targets. The model performs better than the best reported results on Challenge \cite{jiang-etal-2019-challenge}, and similar to previous results on the SemEval corpora. Regarding the choice of target representation, \first is the strongest overall, with an average of 64.7 \F across the original eight datasets, followed by \maxm (64.6), \mean (64.4), \maxmm (64.2), and finally \cls (64.1). It is, however, unclear exactly which representation is the best, as it differs for each dataset. But we can conclude that \cls is in general the weakest model, while either \first or \maxm provide good starting points.

Adding holder annotations to the input text delivers only small improvements on four of the fifteen experiments, and has losses on seven. The +exp. model, however, leads to significant improvements on 10 experiments. The outlier seems to be Darmstadt Services, which contains a large number of ``polar targets'' in the data, which do not have polar expressions. This may explain why including this information has less effect on this dataset. Finally, +full performs between the original input and +exp.

\subsection{Polarity classification with predicted annotations}

The expression models achieve modest \F scores when trained and tested on the same dataset -- between 15.0 and 47.9 --, and poor scores when transferred to a different dataset -- between 0.9 and 14.9 (further details shown in Table \ref{tab:expresults} in Appendix \ref{app:additional}). The lexicons often provide better cross-dataset \F than the expression models trained on another dataset, as they have relatively good precision on general sentiment terms.

\begin{figure}
    \centering
    \includegraphics[width=.5\textwidth]{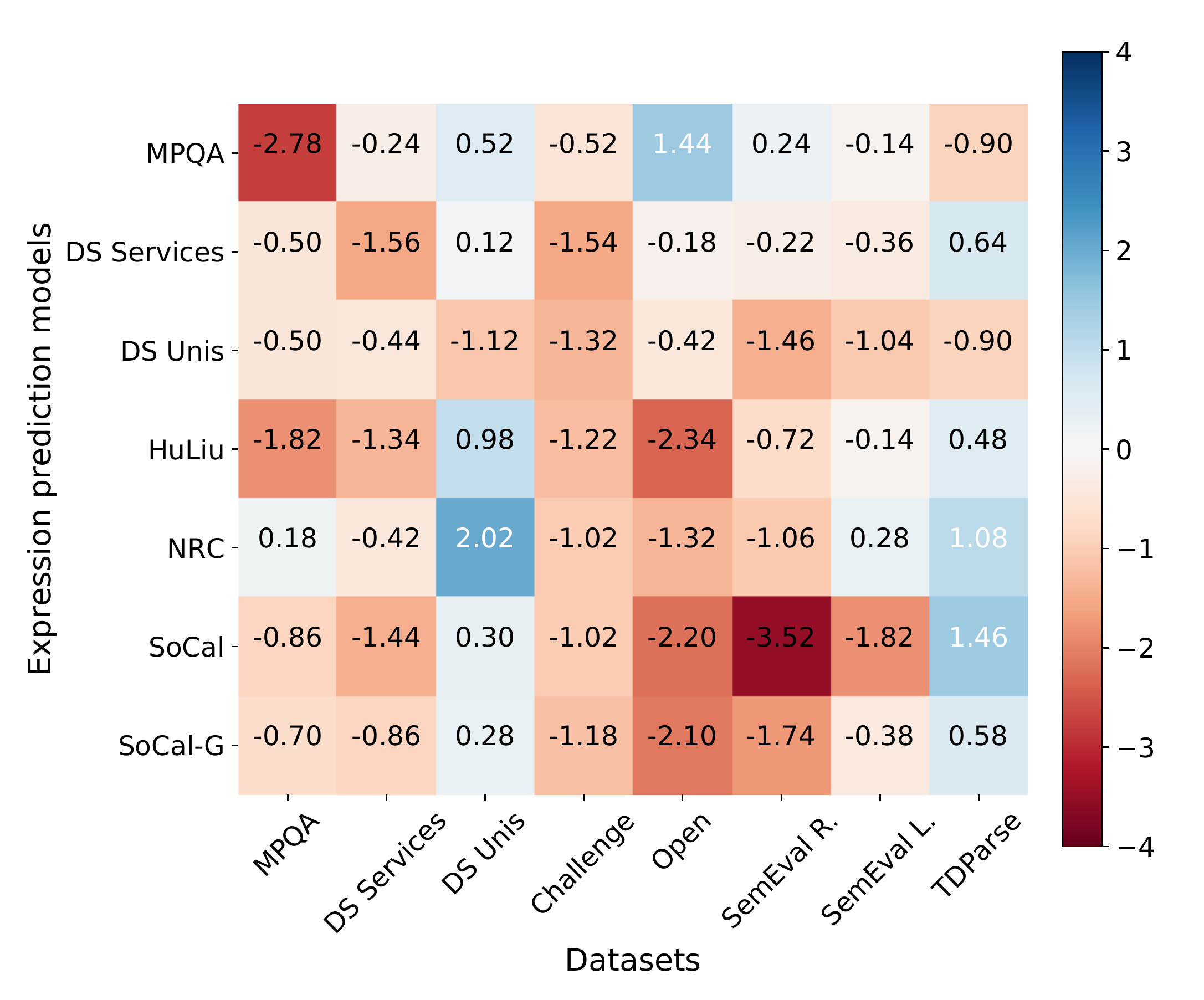}
    \caption{Heatmap of average improvements (blue) and losses (red) on the target classification tasks (x-axis) when augmenting the input text with predicted sentiment expressions from the expression prediction models (y-axis).}
    \label{fig:heatmap}
\end{figure}

Figure \ref{fig:heatmap} shows a heatmap of improvements (blue) and losses (red) on the eight datasets (x-axis) when augmenting the input text with expression tags from the expression models and lexicons (y-axis). We compare the expression augmented results to the original results for each pooling technique and take the average of these improvements and losses. For a full table of all results, see Table \ref{tab:predicted_sent_results} in Appendix \ref{app:additional}.

Augmenting the input text with predicted sentiment expressions leads to losses in 41 out of averaged 56 experiments shown in Figure \ref{fig:heatmap} (or in 173 out of 280 experiments in Table \ref{tab:predicted_sent_results}). Curiously, the experiments that use an expression model trained on the same dataset as the classification task, \eg MPQA predicted expressions on the MPQA classification task, have the largest losses -- the largest of which is MPQA (-2.78 on average). This seems to indicate that the mismatch between the train prediction, which are near perfect, and the rather poor test predictions is more problematic than cross-dataset predictions, which are similar on train and test.

\begin{table*}[]

\newcommand{\sep}{\cmidrule(lr){3-3}\cmidrule(lr){4-4}\cmidrule(lr){5-5}\cmidrule(lr){6-6}\cmidrule(lr){7-7}\cmidrule(lr){8-8}\cmidrule(lr){9-9}\cmidrule(lr){10-10}}
    \centering
    \resizebox{\textwidth}{!}{%
    \begin{tabular}{llrrrrrrrrr}
    \toprule
    & & MPQA & DS. Services & DS. Unis & Challenge & SemEval R. & SemEval L. & Open & TDParse \\
    \sep
 & Previous Results & n/a  & n/a & n/a & 70.3 & 80.1 & 78.3 & &\\

\midrule
\multirow{4}{*}{\rotatebox{90}{\cls}}
 & original  &   63.5 \f{2} & 57.3 \f{1} &  57.6 \f{4} & 84.3 \f{0} & 74.1 \f{2} & 72.8 \f{1} & 54.6 \f{1} & 48.8 \f{1} \\
& + holders             &  \loss{63.1 \f{2}} &  57.1 \f{1} & \imp{60.5 \f{0}} &  - & - & - & - & -\\
& + exp.                &  \imp{64.0  \f{3}} & \loss{56.4 \f{0}} &  \imp{62.9 \f{4}} &  - & - & - & - & -\\
& + full     &  \loss{61.9  \f{2}} & \loss{56.6 \f{1}} &  \imp{62.8 \f{2}} &   - & - & - & - & -\\

\midrule

\multirow{4}{*}{\rotatebox{90}{\first}}
& original &  64.3 \f{2} & 57.8 \f{1} &  58.7 \f{4} & \textbf{84.4 \f{1}} & 75.6 \f{1} & 74.3 \f{1} & 55.6 \f{2} & 46.6 \f{1} \\
& + holders            &  \loss{63.4  \f{2} }&  57.7 \f{2} &  \imp{60.5  \f{3}} &  - & - & - & - & -\\
& + exp.               &  \imp{\textbf{64.8  \f{2}}} &  \loss{57.0 \f{1}} & \imp{63.7  \f{2}} &  - & - & - & - & -\\
& + full    &  64.0  \f{1} &  \loss{55.2 \f{1}} &  \imp{\textbf{65.7  \f{4}}} &  - & - & - & - & -\\

\midrule

\multirow{4}{*}{\rotatebox{90}{\mean}}
& original  & 63.5 \f{2}  & 57.3 \f{1} &  60.2 \f{4} & \textbf{84.4 \f{1}} & 74.1 \f{2} & 72.8 \f{1} & 56.8 \f{3} & 46.1 \f{1}  \\
& + holders &  \loss{63.1 \f{2}} &   \imp{57.8  \f{1}} & \loss{56.7  \f{5}} &  - & - & - & - & -\\
& + exp.   &  \imp{64.3 \f{2}} &   \loss{56.2  \f{1}} & \imp{64.1  \f{3}} &  - & - & - & - & -\\
& + full    &  \imp{64.2 \f{2}} &   \loss{56.3  \f{1}} & \imp{63.7  \f{2}} &  - & - & - & - & -\\

\midrule

\multirow{4}{*}{\rotatebox{90}{\maxm}}
& original        &  60.8 \f{4}  & 58.2 \f{1} &  57.8 \f{3} & 81.4 \f{1} & 73.9 \f{2} & \textbf{74.5 \f{2}} & \textbf{61.4 \f{5}} & \textbf{49.0 \f{3}}  \\
& + holders      &  \loss{61.9 \f{4}} & 57.9 \f{1} &  \loss{53.9 \f{1}} & - & - & - & - & -\\
& + exp.      &  \imp{64.3 \f{2}} &  \loss{57.4 \f{1}} & \imp{61.5 \f{6}} & - & - & - & - & -\\
& + full  &  \imp{62.7 \f{3}} & 57.9 \f{1} &  \loss{54.5  \f{2}} & - & - & - & - & -\\

\midrule

\multirow{4}{*}{\rotatebox{90}{\maxmm}}
& original  &  59.3 \f{2} & 57.8 \f{1} &  55.2 \f{3} & 81.3 \f{1} & \textbf{77.2 \f{1}} & \textbf{74.5 \f{1}} & 60.2 \f{5} & 48.5 \f{5} \\
& + holders &  \imp{61.3 \f{1}} & 57.8 \f{1} & \loss{54.7 \f{3}} & - & - & - & - & -\\
& + exp.    &  \imp{64.1 \f{2}} & \imp{\textbf{59.8 \f{3}}} & \loss{54.0 \f{2}} &  - & - & - & - & -\\
& + full  &  \imp{63.9 \f{1}} & 57.7 \f{1} & \loss{54.4 \f{4}} &  - & - & - & - & -\\

       \bottomrule
    \end{tabular}
}%
    
    \caption{Average macro \F scores for \underline{polarity classification} across five runs (standard deviation in parenthesis) on gold targets, also adding information about holders and expressions. \textbf{Bold} indicates the best model per dataset, while \imp{blue} and \loss{pink} highlighting indicates an \imp{improvement} or \loss{loss in performance} compared to the original (targets only) data, respectively.}
    \label{tab:sentimentresults}
\end{table*}

The best expression prediction model is the one trained on MPQA, improving the performance on Darmstadt Universties, Open, and SemEval Restaurants. This is likely due to the fact that MPQA has the largest number of annotated expressions, and that the domain is more general, leading to expression predictions that generalize better. The expression models trained on Darmstadt Services leads to small benefits on two corpora and the expression model trained on Darmstadt Universities only leads to losses

The datasets that receive the most benefit from expression annotations are Darmstadt Universities (6/7 experiments) and the TDParse dataset (5/7). In both cases, the lexicon-based expression models provide more consistent benefits than the trained expression prediction models. The fact that the dataset that benefits most is the TDParse dataset suggests that expression information is most useful when there are multiple targets with multiple polarities.

 There is no significant correlation between the performance of the expression prediction model and the performance on the classification task on the three fine-grained datasets. In fact, there is a small but insignificant negative correlation (-0.33 p$=$0.13, -0.16 p$=$0.48, -0.26 p$=$0.25 for macro Precision, Recall, or \F respectively, as measured by Pearson's correlation between the expression performances and the \F of the classification models augmented with these predicted expressions). It seems that the possible benefits depends more on the target dataset than the actual expression model used.

\section{Conclusion}
\label{sec:summary}

In this work we have explored the benefit of augmenting targeted sentiment models with holder and sentiment expressions. The experiments have shown that although augmenting text with holder and expression tags (\textbf{RQ1~a}) or simultaneously predicting them (\textbf{RQ1~b}) have no benefit for target extraction, predicting collapsed BIO + polarity tags consistently improves target extraction (\textbf{RQ1~c}). Furthermore, augmenting the input text with gold expressions generally improves targeted polarity classification (\textbf{RQ2~a}), although it is not clear which target representation strategy is best (\textbf{RQ2~b}). Furthermore, we have found benefits of including lexicon-based expressions for the more complex targeted datasets (\textbf{RQ2~c}).

The rather poor performance of the learned expression models and the difference between augmenting with gold or predicted expressions reveals the need to improve expression prediction approaches, both by creating larger corpora annotated with sentiment expressions, as well as performing further research on the modeling aspect. Any future work interested in modelling more complex sentiment phenomena should therefore be aware that we may first require more high-quality annotated data if we wish to do so with current state-of-the-art machine learning approaches.

Furthermore, we introduce a common format for eight standard English datasets in fine-grained sentiment analysis and release the scripts to download and preprocess them easily. We plan to include further datasets in our script in the future, as well as extending our work to other languages with available fine-grained corpora.

%

\bibliographystyle{acl_natbib}
\interlinepenalty=10000
%
\bibliography{lit}

\begin{thebibliography}{23}
\expandafter\ifx\csname natexlab\endcsname\relax\def\natexlab#1{#1}\fi

\bibitem[{Devlin et~al.(2019)Devlin, Chang, Lee, and
  Toutanova}]{devlin-etal-2019-bert}
Jacob Devlin, Ming-Wei Chang, Kenton Lee, and Kristina Toutanova. 2019.
\newblock \href {https://doi.org/10.18653/v1/N19-1423} {{BERT}: Pre-training of
  deep bidirectional transformers for language understanding}.
\newblock In \emph{Proceedings of the 2019 Conference of the North {A}merican
  Chapter of the Association for Computational Linguistics: Human Language
  Technologies, Volume 1 (Long and Short Papers)}, pages 4171--4186,
  Minneapolis, Minnesota. Association for Computational Linguistics.

\bibitem[{Etter et~al.(2013)Etter, Ferraro, Cotterell, Buzek, and {Van
  Durme}}]{nerit-named-entity-recognition-for-informal-text}
David Etter, Francis Ferraro, Ryan Cotterell, Olivia Buzek, and Benjamin {Van
  Durme}. 2013.
\newblock \href {http://www.cs.jhu.edu/~ferraro/papers/etter-nerit-2013.pdf}
  {{Nerit: Named Entity Recognition for Informal Text}}.
\newblock Technical Report~11, Human Language Technology Center of Excellence,
  Johns Hopkins University, Baltimore, Maryland.

\bibitem[{He et~al.(2019)He, Lee, Ng, and Dahlmeier}]{he-etal-2019-interactive}
Ruidan He, Wee~Sun Lee, Hwee~Tou Ng, and Daniel Dahlmeier. 2019.
\newblock \href {https://doi.org/10.18653/v1/P19-1048} {An interactive
  multi-task learning network for end-to-end aspect-based sentiment analysis}.
\newblock In \emph{Proceedings of the 57th Annual Meeting of the Association
  for Computational Linguistics}, pages 504--515, Florence, Italy. Association
  for Computational Linguistics.

\bibitem[{He and McAuley(2016)}]{He2016-ups-downs}
Ruining He and Julian McAuley. 2016.
\newblock \href {https://doi.org/10.1145/2872427.2883037} {Ups and downs:
  Modeling the visual evolution of fashion trends with one-class collaborative
  filtering}.
\newblock In \emph{Proceedings of the 25th International Conference on World
  Wide Web}, WWW ’16, page 507–517, Republic and Canton of Geneva, CHE.
  International World Wide Web Conferences Steering Committee.

\bibitem[{Hu et~al.(2019)Hu, Peng, Huang, Li, and Lv}]{hu-etal-2019-open}
Minghao Hu, Yuxing Peng, Zhen Huang, Dongsheng Li, and Yiwei Lv. 2019.
\newblock \href {https://doi.org/10.18653/v1/P19-1051} {Open-domain targeted
  sentiment analysis via span-based extraction and classification}.
\newblock In \emph{Proceedings of the 57th Annual Meeting of the Association
  for Computational Linguistics}, pages 537--546, Florence, Italy. Association
  for Computational Linguistics.

\bibitem[{Hu and Liu(2004)}]{hu2004mining}
Minqing Hu and Bing Liu. 2004.
\newblock Mining and summarizing customer reviews.
\newblock In \emph{Proceedings of the 10th ACM SIGKDD International Conference
  on Knowledge Discovery and Data Mining}, pages 168--177, Seattle, USA.

\bibitem[{Jiang et~al.(2019)Jiang, Chen, Xu, Ao, and
  Yang}]{jiang-etal-2019-challenge}
Qingnan Jiang, Lei Chen, Ruifeng Xu, Xiang Ao, and Min Yang. 2019.
\newblock \href {https://doi.org/10.18653/v1/D19-1654} {A challenge dataset and
  effective models for aspect-based sentiment analysis}.
\newblock In \emph{Proceedings of the 2019 Conference on Empirical Methods in
  Natural Language Processing and the 9th International Joint Conference on
  Natural Language Processing (EMNLP-IJCNLP)}, pages 6279--6284, Hong Kong,
  China. Association for Computational Linguistics.

\bibitem[{Katiyar and Cardie(2016)}]{katiyar-cardie-2016-investigating}
Arzoo Katiyar and Claire Cardie. 2016.
\newblock \href {https://doi.org/10.18653/v1/P16-1087} {Investigating {LSTM}s
  for joint extraction of opinion entities and relations}.
\newblock In \emph{Proceedings of the 54th Annual Meeting of the Association
  for Computational Linguistics (Volume 1: Long Papers)}, pages 919--929,
  Berlin, Germany. Association for Computational Linguistics.

\bibitem[{Klinger and Cimiano(2013)}]{klinger-cimiano-2013-bi}
Roman Klinger and Philipp Cimiano. 2013.
\newblock \href {https://www.aclweb.org/anthology/P13-2147} {Bi-directional
  inter-dependencies of subjective expressions and targets and their value for
  a joint model}.
\newblock In \emph{Proceedings of the 51st Annual Meeting of the Association
  for Computational Linguistics (Volume 2: Short Papers)}, pages 848--854,
  Sofia, Bulgaria. Association for Computational Linguistics.

\bibitem[{Li et~al.(2019)Li, Bing, Li, and Lam}]{li-etal-2019-unified}
Xin Li, Lidong Bing, Piji Li, and Wai Lam. 2019.
\newblock \href {https://doi.org/10.1609/aaai.v33i01.33016714} {A unified model
  for opinion target extraction and target sentiment prediction}.
\newblock In \emph{Proceedings the Thirty-Third AAAI Conference on Artificial
  Intelligence (AAAI 2019)}, pages 6714--6721, Honolulu, Hawaii. AAAI Press.

\bibitem[{Liu(2015)}]{Liu:15}
Bing Liu. 2015.
\newblock \emph{Sentiment analysis: Mining Opinions, Sentiments, and Emotions}.
\newblock Cambridge University Press, Cambridge, United Kingdom.

\bibitem[{Loper and Bird(2002)}]{Loper2002}
Edward Loper and Steven Bird. 2002.
\newblock {NLTK: The natural language toolkit}.
\newblock In \emph{Proceedings of the ACL-02 Workshop on Effective Tools and
  Methodologies for Teaching Natural Language Processing and Computational
  Linguistics - Volume 1}, pages 63--70.

\bibitem[{Mitchell et~al.(2013)Mitchell, Aguilar, Wilson, and
  Van~Durme}]{mitchell-etal-2013-open}
Margaret Mitchell, Jacqui Aguilar, Theresa Wilson, and Benjamin Van~Durme.
  2013.
\newblock \href {https://www.aclweb.org/anthology/D13-1171} {Open domain
  targeted sentiment}.
\newblock In \emph{Proceedings of the 2013 Conference on Empirical Methods in
  Natural Language Processing}, pages 1643--1654, Seattle, Washington, USA.
  Association for Computational Linguistics.

\bibitem[{Mohammad and Turney(2013)}]{Mohammad13}
Saif~M. Mohammad and Peter~D. Turney. 2013.
\newblock {Crowdsourcing a Word-Emotion Association Lexicon}.
\newblock \emph{Computational Intelligence}, 29(3):436--465.

\bibitem[{Pontiki et~al.(2014)Pontiki, Galanis, Pavlopoulos, Papageorgiou,
  Androutsopoulos, and Manandhar}]{pontiki-etal-2014-semeval}
Maria Pontiki, Dimitris Galanis, John Pavlopoulos, Harris Papageorgiou, Ion
  Androutsopoulos, and Suresh Manandhar. 2014.
\newblock \href {https://doi.org/10.3115/v1/S14-2004} {{S}em{E}val-2014 task 4:
  Aspect based sentiment analysis}.
\newblock In \emph{Proceedings of the 8th International Workshop on Semantic
  Evaluation ({S}em{E}val 2014)}, pages 27--35, Dublin, Ireland. Association
  for Computational Linguistics.

\bibitem[{Rietzler et~al.(2020)Rietzler, Stabinger, Opitz, and
  Engl}]{rietzler-etal-2020-adapt}
Alexander Rietzler, Sebastian Stabinger, Paul Opitz, and Stefan Engl. 2020.
\newblock \href {https://www.aclweb.org/anthology/2020.lrec-1.607} {Adapt or
  get left behind: Domain adaptation through {BERT} language model finetuning
  for aspect-target sentiment classification}.
\newblock In \emph{Proceedings of The 12th Language Resources and Evaluation
  Conference}, pages 4933--4941, Marseille, France. European Language Resources
  Association.

\bibitem[{Taboada et~al.(2006)Taboada, Anthony, and
  Voll}]{taboada-etal-2006-methods}
Maite Taboada, Caroline Anthony, and Kimberly Voll. 2006.
\newblock \href {http://www.lrec-conf.org/proceedings/lrec2006/pdf/420_pdf.pdf}
  {{Methods for Creating Semantic Orientation Dictionaries}}.
\newblock In \emph{Proceedings of the Fifth International Conference on
  Language Resources and Evaluation ({LREC}{'}06)}, Genoa, Italy. European
  Language Resources Association (ELRA).

\bibitem[{Tang et~al.(2014)Tang, Wei, Yang, Zhou, Liu, and
  Qin}]{tang-etal-2014-learning}
Duyu Tang, Furu Wei, Nan Yang, Ming Zhou, Ting Liu, and Bing Qin. 2014.
\newblock \href {https://doi.org/10.3115/v1/P14-1146} {Learning
  sentiment-specific word embedding for twitter sentiment classification}.
\newblock In \emph{Proceedings of the 52nd Annual Meeting of the Association
  for Computational Linguistics (Volume 1: Long Papers)}, pages 1555--1565,
  Baltimore, Maryland. Association for Computational Linguistics.

\bibitem[{Toprak et~al.(2010)Toprak, Jakob, and
  Gurevych}]{toprak-etal-2010-sentence}
Cigdem Toprak, Niklas Jakob, and Iryna Gurevych. 2010.
\newblock \href {https://www.aclweb.org/anthology/P10-1059} {Sentence and
  expression level annotation of opinions in user-generated discourse}.
\newblock In \emph{Proceedings of the 48th Annual Meeting of the Association
  for Computational Linguistics}, pages 575--584, Uppsala, Sweden. Association
  for Computational Linguistics.

\bibitem[{Wang et~al.(2017)Wang, Liakata, Zubiaga, and
  Procter}]{wang-etal-2017-tdparse}
Bo~Wang, Maria Liakata, Arkaitz Zubiaga, and Rob Procter. 2017.
\newblock \href {https://www.aclweb.org/anthology/E17-1046} {{TDP}arse:
  Multi-target-specific sentiment recognition on twitter}.
\newblock In \emph{Proceedings of the 15th Conference of the {E}uropean Chapter
  of the Association for Computational Linguistics: Volume 1, Long Papers},
  pages 483--493, Valencia, Spain. Association for Computational Linguistics.

\bibitem[{Wiebe et~al.(2005)Wiebe, Wilson, and Cardie}]{Wiebe2005}
Janyce Wiebe, Theresa Wilson, and Claire Cardie. 2005.
\newblock {Annotating expressions of opinions and emotions in language}.
\newblock \emph{Language Resources and Evaluation (formerly Computers and the
  Humanities)}, 39(2/3):164–210.

\bibitem[{Xu et~al.(2019)Xu, Liu, Shu, and Yu}]{xu-etal-2019-bert}
Hu~Xu, Bing Liu, Lei Shu, and Philip Yu. 2019.
\newblock \href {https://doi.org/10.18653/v1/N19-1242} {{BERT} post-training
  for review reading comprehension and aspect-based sentiment analysis}.
\newblock In \emph{Proceedings of the 2019 Conference of the North {A}merican
  Chapter of the Association for Computational Linguistics: Human Language
  Technologies, Volume 1 (Long and Short Papers)}, pages 2324--2335,
  Minneapolis, Minnesota. Association for Computational Linguistics.

\bibitem[{Zhang et~al.(2015)Zhang, Zhang, and Vo}]{zhang-etal-2015-neural}
Meishan Zhang, Yue Zhang, and Duy-Tin Vo. 2015.
\newblock \href {https://doi.org/10.18653/v1/D15-1073} {Neural networks for
  open domain targeted sentiment}.
\newblock In \emph{Proceedings of the 2015 Conference on Empirical Methods in
  Natural Language Processing}, pages 612--621, Lisbon, Portugal. Association
  for Computational Linguistics.

\end{thebibliography}

\clearpage
\appendices

\section{Additional tables}
\label{app:additional}

\begin{table*}[h!]
    \centering

    \newcommand{\sep}{\cmidrule(lr){4-4}\cmidrule(lr){5-5}\cmidrule(lr){6-6}\cmidrule(lr){7-7}\cmidrule(lr){8-8}\cmidrule(lr){9-9}\cmidrule(lr){10-10}\cmidrule(lr){11-11}}
    \centering
    \resizebox{\textwidth}{!}{%
    \begin{tabular}{lllrrrrrrrrr}
    \toprule
    
  &  & & MPQA & DS. Services & DS. Unis & Challenge & Open & SemEval R. & SemEval L.  & TDParse \\
    \sep

& \multirow{5}{*}{\rotatebox{90}{original}} 
& \cls   & 63.5 \f{2} & 57.3 \f{1} &  57.6 \f{4} & 84.3 \f{0} & 54.6 \f{1}  &74.1 \f{2} & 72.8 \f{1} & 48.8 \f{1} \\
& & \first & \underline{\textbf{64.3 \f{2}}} & 57.8 \f{1} &  58.7 \f{4} & \textbf{\underline{84.4 \f{1}}} & 55.6 \f{2} & 75.6 \f{1} & 74.3 \f{1} & 46.6 \f{1} \\
& & \mean & 63.5 \f{2}  & 57.3 \f{1} &  60.2 \f{4} & \textbf{\underline{84.4 \f{1}}} & 56.8 \f{3} & 74.1 \f{2} & 72.8 \f{1} & 46.1 \f{1}  \\
& & \maxm &  60.8 \f{4}  & \underline{\textbf{58.2 \f{1}}} &  57.8 \f{3} & 81.4 \f{1} & 61.4 \f{5} & 73.9 \f{2} & 74.5 \f{2}  & 49.0 \f{3}  \\
& & \maxmm & 59.3 \f{2} & 57.8 \f{1} &  55.2 \f{3} & 81.3 \f{1} & 60.2 \f{5} & \underline{\textbf{77.2 \f{1}}} & 74.5 \f{1} & 48.5 \f{5} \\

\midrule

\multirow{17}{*}{\rotatebox{90}{Predicted Expressions}}

& \multirow{5}{*}{\rotatebox{90}{MPQA}} 
& \cls   & \loss{60.3 \f{2}} & \loss{57.0 \f{1}} & \imp{61.3 \f{5}} & \loss{83.1 \f{1}} & \imp{57.5 \f{4}} & 74.2 \f{2} & \loss{72.2 \f{1}} & \loss{47.5 \f{2}} \\
& & \first & \loss{61.6 \f{2}} & \loss{57.0 \f{2}} & \imp{59.8 \f{3}} & \loss{83.5 \f{1}} & \loss{55.2 \f{3}} & \imp{77.1 \f{1}} & \loss{73.9 \f{2}} & \loss{45.2 \f{1}} \\
& & \mean  & \loss{60.3 \f{2}} & \loss{57.0 \f{1}} & \imp{61.3 \f{5}} & \loss{83.1 \f{1}} & \imp{57.5 \f{4}} & 74.2 \f{2} & \loss{72.2 \f{1}} & \imp{47.5 \f{2}} \\
& & \maxm   & \loss{59.1 \f{2}} & 58.1 \f{0} & \loss{57.0 \f{4}} & \imp{82.3 \f{0}} & \underline{\textbf{\imp{63.7 \f{1}}}} & \imp{75.0 \f{2}} & 74.7 \f{1} & \loss{48.5 \f{2}} \\
& & \maxmm & \loss{56.2 \f{5}} & \imp{58.1 \f{1}} & \loss{52.7 \f{2}} & 81.3 \f{1} & \imp{61.9 \f{3}} & \loss{75.6 \f{2}} & \imp{75.2 \f{1}} & \loss{45.8 \f{4}} \\

\sep

& \multirow{5}{*}{\rotatebox{90}{DS. Services}} 
& \cls  & 63.6 \f{1} &  \loss{56.3 \f{1}} &  \imp{60.6 \f{1}} & \loss{82.4 \f{1}} & \loss{53.4 \f{4}}& \loss{72.1 \f{2}}& \loss{72.2 \f{1}}& \imp{49.4 \f{2}}\\ 
& & \first& \loss{61.3 \f{2}} &  \loss{54.5 \f{0}} &  \imp{59.4 \f{3}} & \loss{82.6 \f{1}} & \imp{56.2 \f{9}}& \imp{76.3 \f{1}}& \imp{74.8 \f{1}}& \loss{45.4 \f{2}}\\
& & \mean & \imp{64.1 \f{2}} &  \loss{56.0 \f{0}} &  60.1 \f{5} & \loss{82.4 \f{1}}& \loss{56.1 \f{3}}& \loss{73.0 \f{2}}& \loss{72.4 \f{1}}& \imp{49.8 \f{2}}\\
& & \maxm  & \imp{61.4 \f{1}} &  \loss{56.7 \f{1}} &  \loss{55.2 \f{2}} & \loss{80.3 \f{1}} & \imp{63.0 \f{2}} & \imp{76.5 \f{1}} & 74.5 \f{1} & \loss{48.1 \f{3}} \\
& & \maxmm& \loss{58.5 \f{2}} &  \loss{57.1 \f{1}} &  \loss{54.8 \f{5}} & \loss{80.4 \f{1}} & \loss{59.0 \f{2}} & \loss{75.9 \f{2}} & \loss{73.2 \f{2}} & \imp{49.5 \f{4}} \\

\sep

& \multirow{5}{*}{\rotatebox{90}{DS. Unis}} 
& \cls   &\loss{63.1 \f{1}}& \loss{57.0 \f{1}}& \imp{60.3 \f{3}}& \loss{82.8 \f{1}} & 54.7 \f{2}& \loss{73.6 \f{3}}& 72.8 \f{1}& \loss{48.1 \f{3}} \\
& & \first &64.1 \f{2}& \loss{56.9 \f{1}}& \loss{58.2 \f{2}}& \loss{82.6 \f{1}} & \loss{55.2 \f{3}}& \loss{70.8 \f{3}}& \loss{72.9 \f{2}}& \loss{44.4 \f{1}} \\
& & \mean  &\loss{62.3 \f{1}}& \loss{57.0 \f{1}}& \loss{59.7 \f{3}}& \loss{82.8 \f{1}} & \loss{54.7 \f{2}}& \loss{73.6 \f{3}}& 72.8 \f{1}& \imp{48.1 \f{3}}  \\
& & \maxm   &\loss{59.6 \f{4}}& \loss{57.8 \f{1}}& \loss{53.4 \f{1}}& \loss{80.5 \f{1}} & \imp{62.0 \f{1}}& \imp{74.8 \f{2}}& \loss{71.0 \f{1}}& \loss{46.9 \f{2}} \\
& & \maxmm &\imp{59.8 \f{3}}& \loss{57.5 \f{1}}& \loss{52.3 \f{4}}& \loss{80.5 \f{0}} & 59.9 \f{4}& \loss{74.8 \f{1}}& \loss{74.2 \f{1}}& \loss{47.0 \f{4}} \\

\midrule

\multirow{20}{*}{\rotatebox{90}{Lexicon Expressions}}
& \multirow{5}{*}{\rotatebox{90}{HuLiu}} 
& \cls   &\loss{ 60.3 \f{2}}& \loss{56.2 \f{1}}& \imp{60.8 \f{3}}& \loss{82.8 \f{1}}& \loss{54.0 \f{3}}& \loss{73.6 \f{1}}& \imp{73.2 \f{1}}& \underline{\textbf{\imp{50.6 \f{1}}}}\\
& & \first & \loss{61.2 \f{2}} & \loss{55.0 \f{1}} & \imp{61.2 \f{1}} & \loss{82.3 \f{1}} & \loss{46.5 \f{3}} & \loss{74.8 \f{1}} & 74.2 \f{1} & \loss{43.8 \f{1}} \\
& & \mean  & \loss{60.3 \f{2} }& \loss{56.2 \f{1}} & \imp{60.8 \f{3}} & \loss{82.8 \f{1}} & \loss{54.0 \f{3}} & \loss{73.6 \f{1}} & \imp{73.2 \f{1}} & \underline{\textbf{\imp{50.6 \f{1}}}} \\
& & \maxm   & \loss{59.7 \f{3}} & \loss{57.2 \f{1}} & \loss{56.4 \f{2}} & \loss{81.0 \f{1}} & 61.1 \f{2} & \imp{75.5 \f{2}} & \loss{73.7 \f{3}} & \loss{47.2 \f{3}}  \\
& & \maxmm & \imp{60.8 \f{3}} & \loss{57.1 \f{1}} & 55.2 \f{2} & \loss{80.8 \f{1}} & \imp{61.3 \f{3}} & \loss{73.8 \f{3}} & \loss{73.9 \f{1}} & \imp{49.2 \f{4}} \\
\sep

& \multirow{5}{*}{\rotatebox{90}{NRC}} 
& \cls     & \imp{64.0 \f{2}} & \loss{56.9 \f{1}} & \imp{63.0 \f{2}} & \loss{83.1 \f{1}} & 54.8 \f{3} & \loss{72.0 \f{1}} & 73.0 \f{1} & \imp{49.4 \f{1}} \\
& & \first   & \loss{63.7 \f{2}} & \loss{56.9 \f{1}} & \imp{61.1 \f{2}} & \loss{83.3 \f{1}} & \loss{49.1 \f{5}} & \loss{74.9 \f{3}} & \underline{\textbf{\imp{74.9 \f{0}}}} & \loss{46.1 \f{2}} \\
& & \mean    & \imp{64.0 \f{2}} & \loss{56.9 \f{1}} & \underline{\textbf{\imp{63.0 \f{2}}}} & \loss{83.1 \f{1}} & 54.8 \f{3} & \loss{72.0 \f{1}} & 73.0 \f{1} & \imp{49.4 \f{1}} \\
& & \maxm     & \imp{61.1 \f{3}} & 58.0 \f{1} & \loss{55.6 \f{2}} & \loss{80.4 \f{1}} & \imp{62.0 \f{1}} & \imp{75.3 \f{2}} & 74.6 \f{2} & \imp{49.7 \f{3}} \\
& & \maxmm   & 59.5 \f{3} & 57.6 \f{1} & \imp{56.9 \f{4}} & \loss{80.8 \f{1}} & \imp{61.3 \f{2}} &  \loss{75.4 \f{2}}  & \imp{74.8 \f{1}} & \imp{49.8 \f{4}}\\

\sep

& \multirow{5}{*}{\rotatebox{90}{SoCal}} 
& \cls   & \loss{63.2 \f{2}} & \loss{56.6 \f{1}} & \imp{60.5 \f{4}} & \loss{83.0 \f{0}} & \loss{51.5 \f{4}} & \loss{69.8 \f{1}} & \loss{71.0 \f{1}} & \imp{50.0 \f{1}} \\
& & \first & \loss{61.8 \f{2}} & \loss{53.7 \f{2}} & \imp{59.9 \f{4}} & \loss{81.8 \f{0}} & \loss{51.4 \f{6}} & \loss{72.8 \f{2}} & \loss{73.0 \f{1}} & \loss{45.4 \f{2}} \\
& & \mean  & \loss{63.2 \f{2}} & \loss{56.6 \f{1}} & \imp{60.5 \f{4}} & \loss{83.0 \f{0}} & \loss{51.5 \f{4}} & \loss{69.8 \f{1}} & \loss{71.0 \f{1}} & \imp{50.0 \f{1}} \\
& & \maxm   & \loss{59.2 \f{2}} & \loss{57.8 \f{1}} & \loss{54.5 \f{2}} & \loss{79.3 \f{1}} & \imp{62.3 \f{2}} & \loss{71.5 \f{4}} & \loss{71.8 \f{2}} & \imp{49.5 \f{4}}  \\
& & \maxmm & \imp{59.7 \f{2}} & \imp{56.5 \f{2}} & \imp{55.6 \f{1}} & \loss{79.1 \f{1}} & \imp{60.9 \f{3}} & \loss{73.4 \f{3}} & \loss{73.0 \f{1}} & \imp{51.4 \f{2}}  \\

\sep

& \multirow{5}{*}{\rotatebox{90}{SoCal-Google}} 
& \cls   & \loss{62.6 \f{3}} & \loss{56.5 \f{1}} & \imp{60.0 \f{3}} & \loss{83.0 \f{1}} & \loss{53.2 \f{3}} & \loss{71.5 \f{1}} & 72.9 \f{1} & \imp{50.2 \f{1}} \\
& & \first & \loss{62.1 \f{1}} & \loss{56.2 \f{1}} & \imp{60.8 \f{5}} & \loss{82.5 \f{1}} & \loss{49.7 \f{6}} & \loss{74.9 \f{2}} & 74.0 \f{1} & \loss{46.2 \f{0}} \\
& & \mean  & \loss{62.6 \f{3}} & \loss{56.5 \f{1}} & 60.0 \f{3} & \loss{83.0 \f{1}} & \loss{53.2 \f{3}} & \loss{71.5 \f{1}} & 72.9 \f{1} & \imp{50.2 \f{1}}\\
& & \maxm   & \loss{60.0 \f{3}} & \loss{57.8 \f{0}} & \loss{55.5 \f{5}} & \loss{80.9 \f{1}} & 61.5 \f{3} & \imp{74.8 \f{2}} & 74.5 \f{2} & \imp{49.5 \f{4}} \\
& & \maxmm & \imp{60.6 \f{4}} & \loss{57.1 \f{1}} & \loss{54.6 \f{3}} & \loss{80.5 \f{1}} & \imp{60.5 \f{3}} & \loss{73.5 \f{4}} & \loss{72.7 \f{2}} & \loss{45.8 \f{5}} \\

           \bottomrule
    \end{tabular}
}%
    
        \caption{Macro \F scores for  \underline{polarity classification} of gold targets. \textbf{Bold} numbers indicate the best model per dataset, while \imp{blue} and \loss{pink} highlighting indicates an \imp{improvement} or \loss{loss in performance} compared to the original data (gold targets only), respectively.}
    \label{tab:predicted_sent_results}
\end{table*}

\begin{table*}[]
\newcommand{\sep}{\cmidrule(lr){2-4}\cmidrule(lr){5-6}}
    \centering
    \begin{tabular}{lrrrrr}
    \toprule
            & \multicolumn{3}{c}{\% Unique} & \multicolumn{2}{c}{\% Overlap}\\
                   \sep
        & train & dev & test & train-dev & train-test \\
        \sep
MPQA & \textbf{85.7} & 88.5 & \textbf{89.2} & \underline{15} & \underline{19} \\
DS. Services & 36.2 & 48.6 & 47.5 & 45.0 & 35.6 \\
DS. Uni  & 35.2 & 52.9 & 45.0 & \textbf{58.5} & 47.6\\
TDParse & 33 & 51.8 & 41.7 & 57.4 & 47.3\\
SemEval R. & 36.3 & 59.8 & 49.4 & 56.4 & 33.8\\
SemEval L.  & 45.5 & 71.7 & 64.8 & 48.9 & 33.7\\
Open  & 85 & \textbf{92.4} & 87.1 & 23 & 24\\
Challenge & \underline{23.1} & \underline{39.0} & \underline{39.7} & 54.1 & \textbf{52}\\
    
       \bottomrule
    \end{tabular}
    
    \caption{Analysis of targets in the datasets. \% Unique describes the number of targets that are found only in that split. \% Overlap describes the percentage of dev/test targets that are found in the train set. We disregard partial matches, \eg ``chinese food'' and ``food''.}
    \label{tab:targets}
\end{table*}

\begin{table*}[]
\newcommand{\sep}{\cmidrule(lr){3-3}\cmidrule(lr){4-4}\cmidrule(lr){5-5}}
    \centering
    \begin{tabular}{llrrr}
    \toprule
    && MPQA & DS. Services & DS. Unis  \\
    \sep
    \multirow{3}{*}{\rotatebox{90}{trained}}
   & MPQA        & 15.0 \f{1.7} & 1.0 \f{0.8}  & 2.2 \f{1.2} \\
   & DS. Services& 0.9  \f{0.3} & 47.9 \f{7.3} & 14.9 \f{1.2} \\
   & DS. Unis    & 1.4  \f{0.6} & 10.9 \f{1.5} & 18.5 \f{1.5} \\
    \sep
    \multirow{4}{*}{\rotatebox{90}{lexicons}}
    & HuLiu       & 4.7 & 17.9 & 16.0 \\
    & NRC         & 3.3 & 7.4  & 9.0  \\
    & SoCal       & 2.4 & 13.2 & 13.8 \\
    & SoCal Google& 1.0 & 13.2 & 11.4 \\
    \bottomrule
    \end{tabular}
        \caption{Token-level macro \F scores for expression prediction models (trained) and lexicon expressions (lexicons) when tested on the three fine-grained datasets (x-axis). The trained model scores are the average and standard deviation across five runs with different random seeds. The lexicon models are deterministic and therefore only have a single score.}
    \label{tab:expresults}
\end{table*}

\clearpage
\section{Training details}
\label{app:trainingdetails}

\centering
\begin{tabular}{cc}
\hline
    \textbf{GPU Infrastructure} & 1 NVIDIA P100, 16 GiB RAM \\
\hline
    \textbf{CPU Infrastructure} & Intel Xeon-Gold 6126 2.6 GHz \\
\hline
    \textbf{Number of search trials} & 50 \\
\hline
    \textbf{Domain training duration} & 2580 sec \\
\hline
    \textbf{Extraction fine-tuning duration} & 15381 sec \\
\hline
    \textbf{Classification fine-tuning duration} & 9080 sec \\
\hline
    \textbf{Model implementation} & https://github.com/blinded/for/review \\
\hline
\vspace{0.05cm}
\end{tabular}

\centering
\begin{tabular}{p{0.5\linewidth}p{0.5\linewidth}}
\hline
    \textbf{Hyperparameter} & \textbf{Assignment} \\
\hline
    number of epochs & 50 \\
\hline
    max. sequence length & 128 \\
\hline
    metric early stopping monitored & validation loss \\
\hline
    batch size & 32  \\
\hline
    sentiment dropout & 0.3 \\
\hline
    learning rate optimiser & Bert Adam\\
\hline
    fine-tuning learning rate & 3e-5 \\
\hline    
    learning rate warmup proportion & 0.1 \\
\hline
    regularisation type & L2  \\
\hline
    regularisation value & 0.01 \\
\hline
\end{tabular}

\end{document}